\documentclass[conference]{IEEEtran}

\pdfoutput=1
\IEEEoverridecommandlockouts
\usepackage{cite}
\usepackage{bm}
\usepackage{multirow}
\usepackage{amsmath,amssymb,amsfonts}
\usepackage{graphicx}
\usepackage{textcomp}
\usepackage{xcolor}
\usepackage{booktabs}
\usepackage{subfigure}
\usepackage[linesnumbered,ruled,lined,commentsnumbered]{algorithm2e}
\graphicspath{{figure/}}

\begin{document}

\title{
	Jointly embedding the local and global relations of heterogeneous graph for rumor detection
}

\author{
	\IEEEauthorblockN{Chunyuan Yuan, Qianwen Ma, Wei Zhou\textsuperscript{*}, Jizhong Han, Songlin Hu\textsuperscript{*}}
	\IEEEauthorblockA{
		\textit{School of Cyber Security, University of Chinese Academy of Sciences} \\
		\textit{Institute of Information Engineering, Chinese Academy of Sciences}\\
		Beijing, China \\
		\{yuanchunyuan, maqianwen, zhouwei, hanjizhong, husonglin\}@iie.ac.cn
	}
	\thanks{* is the corresponding author.}
}

\maketitle

\begin{abstract}
	The development of social media has revolutionized the way people communicate, share information and make decisions, but it also provides an ideal platform for publishing and spreading rumors. Existing rumor detection methods focus on finding clues from text content, user profiles, and propagation patterns. However, the local semantic relation and global structural information in the message propagation graph have not been well utilized by previous works. 
	
	In this paper, we present a novel global-local attention network (GLAN) for rumor detection, which jointly encodes the local semantic and global structural information. We first generate a better integrated representation for each source tweet by fusing the semantic information of related retweets with the attention mechanism. Then, we model the global relationships among all source tweets, retweets, and users as a heterogeneous graph to capture the rich structural information for rumor detection. We conduct experiments on three real-world datasets, and the results demonstrate that GLAN significantly outperforms the state-of-the-art models in both rumor detection and early detection scenarios.
\end{abstract}

\section{Introduction}
With the rapid growth of large-scale social media platforms, such as Twitter, and Sina Weibo, rumors on social media have become a major concern. Rumors can propagate very fast and affect the people's choice because of the convenience of social media. However, it is complicated for ordinary people to distinguish rumors from massive amounts of online information, due to the limitation of professional knowledge, time or space. Therefore, it is necessary to develop automatic and assistant approaches to detect rumors at the early stage.

Existing studies on automatically detecting rumors mainly focused on designing effective features from various information sources, including text content~\cite{Castillo_2011, Qazvinian_2011, Popat_2017}, publisher's profiles~\cite{Castillo_2011, Yang_2012} and propagation patterns~\cite{jin2013epidemiological,sampson2016leveraging,ma2017detect}.  However, these feature-based methods are extremely time-consuming, biased, and labor-intensive. Furthermore, if one or several types of hand-crafted features are unavailable, inadequate or manipulated, the effectiveness of these approaches will be affected.


Motivated by the success of deep learning, many recent studies~\cite{liu2018early,song2018ced} apply various neural networks for rumor detection. For example, recurrent neural network~\cite{ma2016detecting} is applied to learn a representation of tweet text over post time. Liu et al.~\cite{liu2018early} modeled the propagation path as multivariate time series, and applied a combination of recurrent and convolutional networks to capture the variations of user characteristics along the propagation path. One major limitation of these approaches is that they ignore the global structural information among different microblogs and users, which however has been shown conducive to provide useful clues for node classification~\cite{tang2015line}.

As is well-known, social media is naturally structured as a heterogeneous graph, with entities such as user, post, geographic location, and hashtag; and relationships such as follower, friendship, retweet, and spatial neighborhood. So, the heterogeneous network provides new and different perspectives of the relationship among microblogs, and thus contains rich information to improve the performance of rumor detection. However, for rumor detection task, most of the previous studies considered that each source microblog is independent and does not affect each other, and thus they did not fully exploit the correlations between the nodes of different types. 

\begin{figure}[htbp]
    \centering
    \includegraphics[scale=0.4]{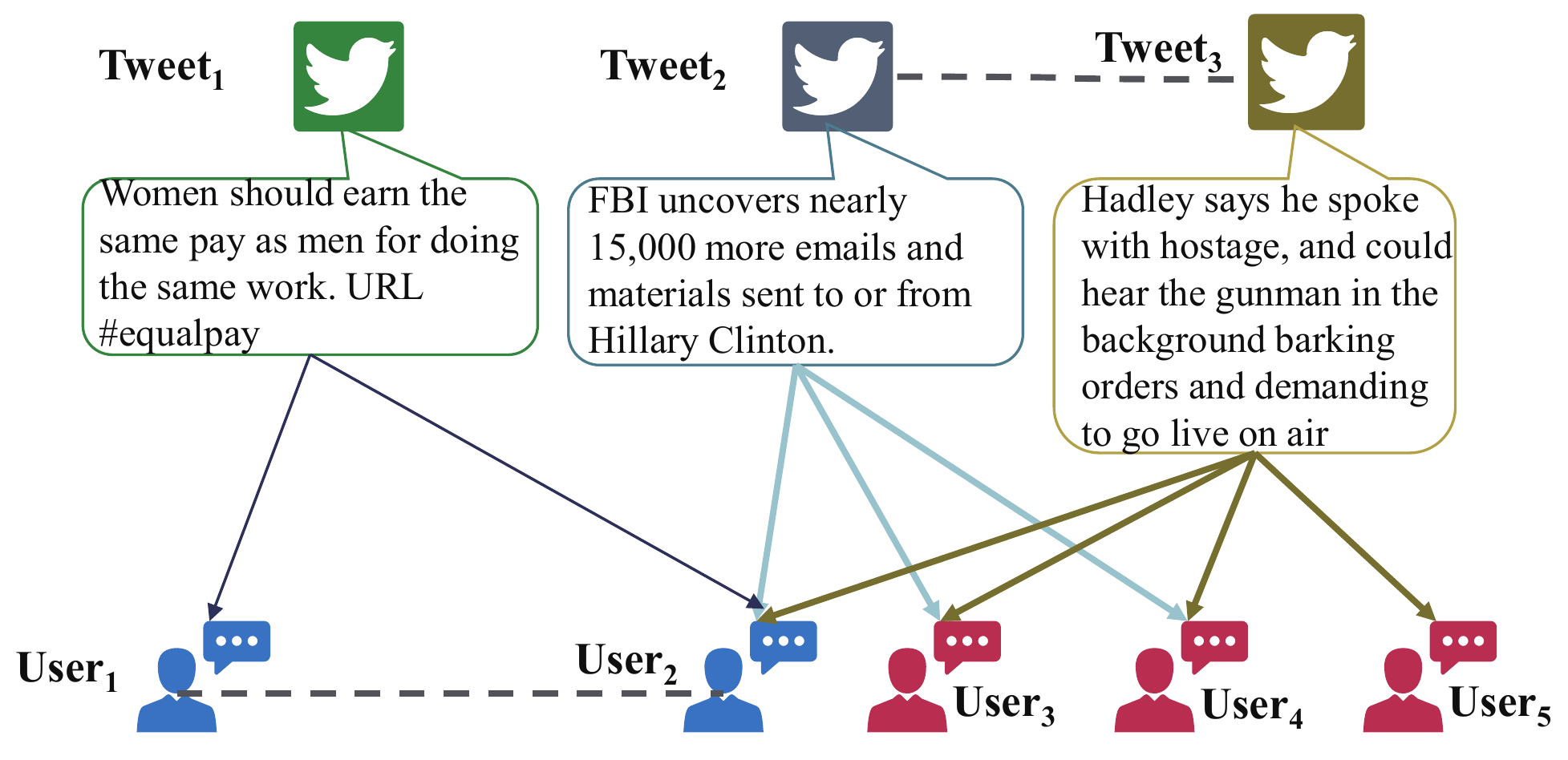}
    \caption{An example of rumor propagation in heterogeneous graph.}
    \label{example}
\end{figure}

To illustrate our motivation, we present a global heterogeneous graph that contains three source tweets with corresponding user responses as shown in Figure 1. For this example, the two users User$_1$ and User$_2$ have no friend relationship, and they do not follow with each other (assumption), however, they all retweeted the same tweet Tweet$_1$. Besides, these three tweets are irrelevant in content, however, the Tweet$_2$ and Tweet$_3$ share similar neighbors, which indicates that they are likely to have the same label. Based on these observations, we construct a global heterogeneous graph to capture the local and global relationships among all source tweets, retweets, and users. Specifically, we connect the user nodes if they participate (tweet or retweet) in common microblogs (e.g., $User_1$--$User_2$) and link the nodes of source tweets by their common users (e.g., $Tweet_2$-$Tweet_3$). In this way, we can learn a latent representation of different type of nodes in the graph. Microblogs and their related users will tend to have close latent representations, so will users publishing the similar tweets or tweets sharing similar participants, even if they are not directly connected in the network.

In this paper, we investigate: (1) how to integrate complex semantics information from learning the retweet sequences; and (2) how to globally model the heterogeneous graph structure of all microblogs and participants for rumor detection. Therefore, to address the two challenges in rumor detection, we propose a novel heterogeneous network with local and global attention for rumor detection. We first aim to fuse the contextual information from source tweet and corresponding retweets with local attention and get a new representation for each source tweet. Then, we construct a global heterogeneous network by combining the different source tweets with structural and semantic properties, instead of detecting rumors from a single microblog.

We evaluate our proposed approach based on three public data sets. The results show that our method outperforms strong rumor detection baselines with a large margin and also demonstrates much higher effectiveness for detection at an early stage of propagation, which is promising for real-time intervention and debunking. 

The contributions of this paper can be summarized as follows:
\begin{itemize}
    \item This is the first study that deeply integrates global structural and local semantic information based on the heterogeneous network for detecting rumors.
    \item We fuse the local contextual information from source tweet and corresponding retweets with multi-head attention and generate a better integrated representation for each source tweet.
    \item We model the global structure as a heterogeneous network by combining the different source tweets with global attention, instead of detecting rumors from a single microblog.
    \item We conduct a series of experiments on three real-world data sets. Experimental results demonstrate that our model achieves superior improvements over state-of-the-art models on both rumor classification and early detection tasks.
\end{itemize}

The rest of the paper is organized as follows. In  Section~\ref{related_works_section}, we briefly review the related work. In Section~\ref{formulation_section}, we formally define the problem of rumor detection with the heterogeneous network. In Section~\ref{model_section}, we introduce the proposed model as well as its training strategy in detail. We conduct experiments on three real-world data sets to evaluate the effectiveness of the proposed model on both rumor classification and early detection tasks in Section~\ref{experiments}. In Section~\ref{parameter_analysis}, we conduct a series of experiments to explore the influence of different hyper-parameters. Finally, we conclude with future work in Section~\ref{conclusion}.

\section{Related Work}  \label{related_works_section}
The target of rumor detection is to distinguish whether a microblog text posted on a social media platform is a rumor or not based on its related information (such as text content, comments, mode of communication, propagation patterns, etc.). Related works can be divided into following categories: (1) Feature-based Classification Methods; (2) Deep Learning Methods; and (3) Propagation Tree Related Methods.


%
%

\subsection{Feature-based Classification Methods} 
Some early studies focus on detecting rumors based on hand-crafted features. These features are mainly extracted from text content and users' profile information. Specifically, Castillo et al. \cite{Castillo_2011} exploited various types of features, i.e., text-based, user-based, topic-based and propagation-based features, to study the credibility of news on Twitter. Yang et al.  Ma et al. \cite{Ma_2015} explored the temporal characteristics of these features based on the time series of rumor's life cycle to incorporate various social context information. Kwon et al. \cite{Kwon_2013} explored a novel approach to identify rumors based on temporal, structural, and linguistic properties of rumor propagation. \cite{Yang_2012} introduced novel features (e.g., the micro-blogging client program used and the event location information) to identify rumors on Sina Weibo. 

%
%
%
%
%

However, the scale and complexity of social media data produce a number of technical challenges. Firstly, the language used in social media is highly informal, ungrammatical, and dynamic, and thus traditional natural language processing techniques cannot be directly applied. Secondly, there are always one or several types of hand-crafted features that are unavailable, inadequate or manipulated. 

\subsection{Deep Learning Methods}
To tackle the above problems of traditional feature-based methods, researchers apply deep learning models to automatically learn efficient features for rumor detection in recent years.  Ma et al.~\cite{ma2016detecting} proposed a recurrent neural networks (RNN) based model to  learn the text representations of relevant posts over time. It is the first study to introduce the deep learning methods into rumor detection. Yu et al.~\cite{Yu_2017} proposed a convolutional method for misinformation identification based on Convolutional Neural Network (CNN), which can capture high-level interactions among significant features. To further improve the detection performance, some studies~\cite{Ruchansky_2017,Bhatt_2018} explore to fuse various features using deep neural networks. For example, the microblog text, the user's profiling data are explored in these studies. Liu et al.~\cite{liu2018early} modeled the propagation path as multivariate time series, and applied both recurrent and convolutional networks to capture the variations of user characteristics along the propagation path.



However, these methods either ignore the propagation patterns or model the propagation path as a sequence structure, which cannot fully utilize the propagation information of the microblog. Furthermore, these methods pay little attention to early detection of rumors. 




\subsection{Propagation Tree Related Methods}
Different from the previous methods that focus on the use of microblog text information, the propagation of trees related methods focuses on the differences in the characteristics of real and false information transmission. 

Jin et al.~\cite{jin2013epidemiological} utilized epidemiological models to characterize information cascades on Twitter resulting from both true news and fake news. Wu et al. \cite{Wu_2015} proposed a random walk graph kernel to model the propagation trees of messages to improve rumor detection.  Sampson et al.~\cite{sampson2016leveraging} applied implicit linkages among conversation fragments about a news story to predict its truthfulness. Ma et al.~\cite{ma2017detect} propose a kernel-based method to capture high-order patterns of microblog posts diffusion with propagation trees, which provide valuable clues on how a microblog is diffused and developed over time.  

However, the message propagation on social media is essentially transmitted in the form of a  heterogeneous graph. In the graph, users delivered or repost a message to make it propagate fast and wide. These methods based on propagation tree only explored the difference in the structure of information transmission, and the relationships among different propagation trees have not been considered. 




\section{Problem Formulation} \label{formulation_section}
Let $\mathcal{M}=\left\{ m_{1},m_{2},\ldots m_{\left| \mathcal{M}\right|} \right\} $ be the set of source microblogs, where each source microblog $m_i$ consists of $n$ retweets $\{r_1, r_2, \ldots, r_n \}$. For every microblog, we use notation $t_i$ to denote their post time. In the heterogeneous graph, we regard retweets as the neighbor nodes of source microblog, which are formulated as $\mathcal{N}(m_i) = [r_1, r_2, \ldots, r_n]$. We denote $\mathcal{U}=\left\{ u_{1},u_{2},\ldots u_{\left| \mathcal{U}\right|} \right\} $ as the set of social media users. 

Our goal is to learn a function $p(c=1| m_i, \mathcal{N}(m_i), \mathcal{U}; \theta)$ to predict whether the source microblog is a rumor or not. $c$ is class label and $\theta$ represents all parameters of the model.

\section{The Proposed Model} \label{model_section}
\begin{figure*}[htbp]
    \centering
    \includegraphics[scale=0.85]{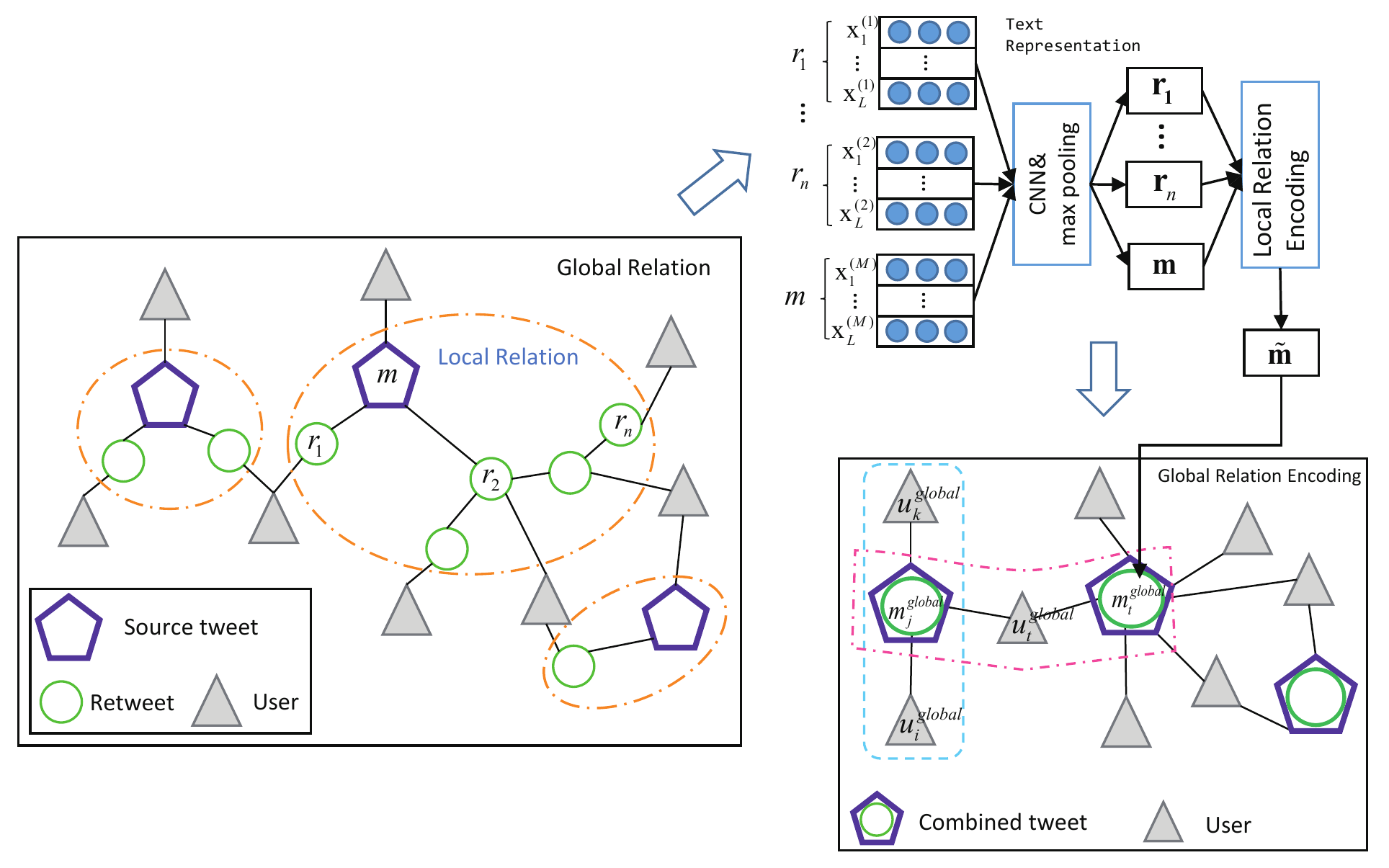}
    \caption{The architecture of the proposed rumor detection model.}
    \label{model}
\end{figure*}

The proposed rumor detection model consists of four major components: microblog representation, local relation encoding,  global relation encoding, and rumor detection with local and global relations. Specifically, the microblog representation module describes the mapping of microblog from word embedding to semantic space; the local relation encoding learns the combined representation of each source tweet from corresponding retweets with multi-head mechanism; the global relation encoding illustrates how to encode the global structure into the node representation; the rumor detection module learns a classification function to predict the label of source microblog. Figure~\ref{model} shows the architecture of the proposed model. Next, we will introduce each of the major components in detail.


\subsection{Microblog Representation}
In this work, the word embedding is used as the representation of a word. We define $\bm{x_j} \in \mathbb{R}^d$ as the $d$-dimensional word embedding corresponding to the $j$-th word in the microblog $\bm{m_i}$. We suppose every microblog has $L$ words. When the length of the microblog is shorter than $L$, zero is padded at the start of the document, and we truncate the microblog at the end position if it is longer than $L$. A sentence of length $L$ is represented as 
\begin{equation}
    \begin{split}
	    & \bm{x}_{1:L}= [\bm{x}_1 ; \bm{x}_2 ; \cdots ; \bm{x}_T] \,,  \\
	\end{split}
\end{equation}
where `;' is the concatenation operator. We use $\bm{x}_{j:j+k}$ to represent the concatenation of words $\bm{x}_j, \bm{x}_{j+1}, \ldots, \bm{x}_{j+k}$.

There have been many neural models to learn the text semantic representation from word sequence embeddings, such as CNN~\cite{kim2014convolutional,kalchbrenner2014convolutional} and RNN~\cite{tai2015improved,yang2016hierarchical}. In this work, we employ the CNN-based model~\cite{kim2014convolutional} as the basic component of the model to learn the semantics of microblogs. 

\subsubsection{CNN}
Given a sequence of words' index $\left(x^{(.)}_1, x^{(.)}_2, \ldots, x^{(.)}_L \right)$ of a microblog, they are transformed to word embeddings $\left( \mathbf{x}^{(.)}_1, \mathbf{x}^{(.)}_2, \ldots, \mathbf{x}^{(.)}_L \right) \in \mathbb{R}^{L \times d} $ by the looking-up layer. Then, convolution layers are applied on word embedding matrix:
\begin{equation}
e_i = \sigma\left(\mathbf{W} * \mathbf{x}^{(.)}_{i:i+h-1} \right) \,,
\end{equation}
to extract feature map $\mathbf{e} = [e_1, e_2, \ldots, e_{L-h+1}] \in \mathbb{R}^{L-h+1}$, where $\mathbf{W} \in \mathbb{R}^{h \times d}$ is the convolutional kernel with $h$ size of receptive filed, and $\sigma(\cdot)$ is non-linear transformation function. Then, we apply a max-overtime pooling operation over the feature map: $ \hat{e} = \max(\mathbf{e})$. 

From above operation, one feature is extracted from one filter. The CNN layer uses $d/3$ filters (with varying receptive filed $h \in \{3,4,5\}$ ) to obtain multiple features. Then, we concatenate all kinds of filters' outputs to form $\mathbf{m}_j \in \mathbb{R}^{d}$ as the representation of the $j$-th microblog $m_j$. By the same way, we can get text representation for every retweet $r_i, i \in [1, n]$. The retweets representation are stack together to form retweet matrix $\mathbf{R} = [\mathbf{r}_1; \mathbf{r}_2; \ldots; \mathbf{r}_n] \in \mathbb{R}^{n \times d}$.

We have obtained microblog representation from word embeddings by the convolutional network. Then, we will introduce how to encode the local relation between source tweet and retweet comments.

\subsection{Local Relation Encoding}
Different from previous studies~\cite{ma2018rumor,liu2018early}, we do not apply recursive neural network or recurrent neural network because they are difficult to parallelize and too costly to use for capturing multi-grained semantic representations. Recently, the attention mechanism shows the superior ability of attention to capture semantic relation, which inspires us to improve semantic representation with an attention mechanism.

\subsubsection{Multi-head Attention}
\begin{figure}[!htp]
	\centering
	\includegraphics[scale=0.9]{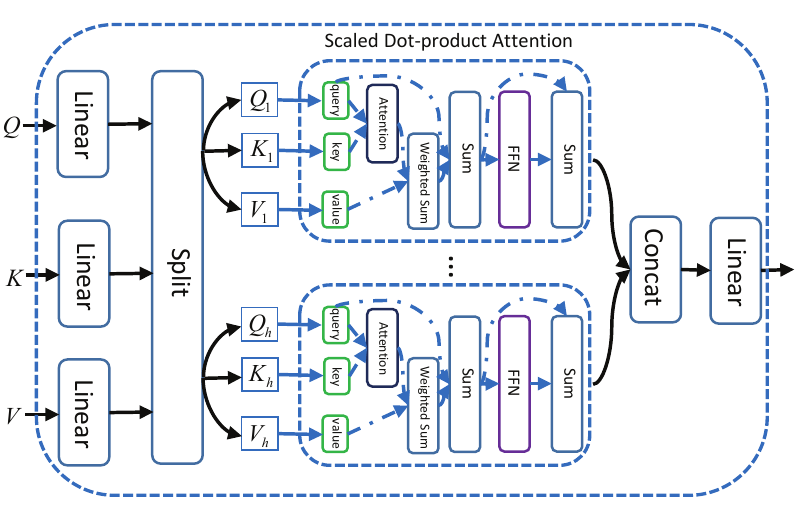}
	\caption{Architecture of Multi-head Attention.}
	\label{attentive_module}
\end{figure}

We use the Multi-head Attention module~\cite{vaswani2017attention} to learn the context information for word representation. Figure~\ref{attentive_module} shows the structure of Multi-head Attention. 

The Multi-head Attention module has three input sentences: the query sentence, the key sentence and the value sentence, namely $Q \in \mathbb{R}^{n_q \times d}$, $K \in \mathbb{R}^{n_k \times d}$, and $V \in \mathbb{R}^{n_v \times d}$ respectively, where $n_q$, $n_k$, and $n_v$ denote the number of words in each sentence, and $d$ is the dimension of the embedding. 

The attention module first takes each word in the query sentence to attend to words in the key sentence via Scaled Dot-Product Attention~\cite{vaswani2017attention}, and then applies those attention weights upon the value sentence:
\begin{equation}
\begin{split}
& \textbf{Attention}(\textbf{Q}, \textbf{K}, \textbf{V}) = \textbf{softmax}\left(\frac{\textbf{Q} \textbf{K}^T}{\sqrt{d}}\right) \textbf{V} \,. \\
\end{split}
\end{equation}

In this way, the attention module can capture dependencies across query sentence and key sentence, and further use the relation information to composite elements in the query sentence and the value sentence into compositional representations.

The entries of $\textbf{V}$ are then linearly combined with the weights to form a new representation of $\textbf{Q}$. In practice, we usually let $\textbf{K} = \textbf{V}$. Thus, a word in $\textbf{Q}$ is represented by its most similar words in $\textbf{V}$. $\textbf{Q}$, $\textbf{K}$, and $\textbf{V}$ are dispensed to $h$ heads. Each head models relationship among $\textbf{Q}$, $\textbf{K}$, and $\textbf{V}$ from one aspect, and corresponds to a scaled dot-product attention module. $\forall i \in [1, h] $ the output of head $i$ is given by
\begin{equation}
\begin{split}
& \mathbf{Z}_i = \textbf{Attention}(\textbf{Q} \mathbf{W}^Q_i, \textbf{K} \mathbf{W}^K_i, \textbf{V} \mathbf{W}^V_i)  \\
\end{split}
\end{equation}
where $\mathbf{W}^Q_i, \mathbf{W}^K_i, \mathbf{W}^V_i \in \mathbb{R}^{d \times d/h}$ are linear transformations. 

Then, the output features of Multi-head Attention are concatenated together and linear transformation is applied to transform it as final output, which is formalized as:
\begin{equation}
\begin{split}
& \mathbf{O} = [\mathbf{Z}_1; \mathbf{Z}_2; \ldots; \mathbf{Z}_h] \mathbf{W}_o \,, \\
\end{split}
\end{equation}
where $\mathbf{W}_o \in \mathbb{R}^{d \times d} $ is a linear transformation. For ease of presentation, we denote a multi-head attention module as $\mathbf{MultiHeadAttention}(Q, K, V)$ and consider two types of representations by varying $Q$, $K$, and $V$.  Note that representations obtained by this module have the same dimension with the query $Q$.

\subsubsection{Local Context Representation} 
To build the connection inside between source microblog and retweet, we first use multi-head attention to refine the representation of every retweet, which is formulated as:
\begin{equation}
\begin{split}
& \mathbf{\widetilde{R}} =  \mathbf{MultiHeadAttention}(\mathbf{R}, \mathbf{R}, \mathbf{R})  \,, \\
\end{split}
\end{equation}
where $\mathbf{\widetilde{R}} \in \mathbb{R}^{n \times d}$. Self-attention lets every retweet attend to each other, and represents each retweet by other similar retweets. As a result, the representation could encode semantic relations among different retweets.

Then, we apply cross attention to build the connection between source tweet and retweet. Specifically, we treat source tweet $\mathbf{m}$ as key and use it to attend the retweet $\mathbf{\widetilde{R}}$ to calculate attention scores for every retweet. 
\begin{equation}
\begin{split}
& \mathbf{s} = \textbf{softmax}(\mathbf{\widetilde{R}} \mathbf{A} \mathbf{m}^T)  \,, \\
& \mathbf{r} = \mathbf{s}^T \mathbf{\widetilde{R}}  \,,
\end{split}
\end{equation}
where $\mathbf{s} \in \mathbb{R}^{n \times 1}$ is an attention score. The score is applied to aggregate the retweets to form a new text representation.

In order to determine the importance between the original microblog representation and the new one, a fusion gate is utilized to combine two representations: 
\begin{equation}
\begin{split}
& \alpha = \sigma(\mathbf{w}_1 \mathbf{m} + \mathbf{w}_2 \mathbf{r} + b)  \,,  \\
& \mathbf{\widetilde{m}} = \mathbf{m} \odot \alpha + \mathbf{r} \odot (1 - \alpha)  \,, 
\end{split}
\end{equation}
where $\sigma(\cdot) = \frac{1}{1+exp(\cdot)}$ is sigmoid activation function, $\mathbf{w}_1, \mathbf{w}_2 \in \mathbb{R}^{d \times 1}$, and $b \in \mathbb{R}$ are learnable parameters of the fusion gate. And $\mathbf{\widetilde{m}}$ is the final microblogs representation.

\subsection{Global Relation Encoding}
After the above local relation encoding process, the heterogeneous graph transforms to the right form of Figure~\ref{model}. In this part, we will discuss how to encode the global structure into the node representation for rumor detection.

We build the global heterogeneous graph which contains two types of nodes: combined text node (source microblog node and retweet node) and user node. Firstly, we clarify the composition of the two types of nodes:
\begin{equation}
    \begin{split}
    & \mathbf{m}^{\prime}_j = \mathbf{m}_j^0 +  \mathbf{\widetilde{m}}_j \,,  \\
    & \mathbf{u}^{\prime}_i = \mathbf{u}_i^0 +  \mathbf{u}_f  \,,
    \end{split}
    \label{global_representation}
\end{equation}
where $\mathbf{m}^0 \in \mathbb{R}^{d}, \mathbf{u}^0 \in \mathbb{R}^{d_u}$ are dynamic vectors which can be updated by the gradient, while $\mathbf{\widetilde{m}}$ and $\mathbf{u}_f$ are static vectors. $\mathbf{u}_f$ is some behavior features or user profile data (such as friends count, followers count, status count, etc.). If $\mathbf{u}_f$ is not available, it can be initialized by a normal distribution.

Considering user node $\mathbf{u}^{\prime}_i$ and text node  $\mathbf{m}^{\prime}_j$ are two different nodes in different semantic space ($d \ne d_u$), we transform them into the same semantic space for further processing, which can be formulated as
\begin{equation}
\begin{split}
& \mathbf{m}^{\prime}_j = \mathbf{W}_m \mathbf{m}^{\prime}_j \,,  \\
& \mathbf{u}^{\prime}_i = \mathbf{W}_u \mathbf{u}^{\prime}_i \,,
\end{split}
\label{transform_representation}
\end{equation}
where $\mathbf{W}_m \in \mathbb{R}^{d \times d}$ and $\mathbf{W}_u \in \mathbb{R}^{d_u \times d}$ are learned parameters.

Recent years, there are many effective methods~\cite{kipf2017semi,hamilton2017inductive,veli2018graph} to encode the homogeneous graph structure into a continuous space. Inspired by the graph attention network~\cite{veli2018graph}, we apply the attention mechanism to learn a distributed representation of each node in the graph by attending over its neighbors.

From the right graph in Figure~\ref{model}, there are two types of relations: (1) user-centric relation (such as $u_k$-$m_j$-$u_i$); and (2) microblog-centric relation (such as $m_j$-$u_t$-$m_t$). To encode these two types of relations into the node representation, we propose to simultaneously attend the neighbor nodes centering on user and microblog node. The attention mechanism is defined as follows:
\begin{equation}
    \begin{split}
    & \alpha^{\prime}_{ji} = \textbf{softmax}(\mathbf{LeakyReLU}( \mathbf{a}^T [\mathbf{m}^{\prime}_j; \mathbf{u}^{\prime}_i] ) ) \,,  \\
    & \beta^{\prime}_{jt} = \textbf{softmax}(\mathbf{LeakyReLU}( \mathbf{c}^T [\mathbf{u}^{\prime}_i; \mathbf{m}^{\prime}_t] ) )   \,,
    \end{split}
    \label{relation_attention}
\end{equation}
where $\mathbf{a}, \mathbf{c} \in \mathbb{R}^{2d \times 1}$ are learned parameters. The parameter of LeakyReLU~\cite{xu2015empirical} non-linearity is set to 0.2.



To capture multiple representations from different relation, we extend our attention mechanism to employ multi-head paradigm, which is similar to Multi-head Attention~\cite{vaswani2017attention}. Specifically, $K$ independent attention mechanisms execute the transformation of Equation~(\ref{relation_attention}), and then their features are concatenated, resulting in the following output feature representation:
\begin{equation}
\begin{split}
& \mathbf{m}^{global}_j = \mathop{||}_{k=1}^{K} \sigma \left( \sum_{i \in \mathcal{N}(m_j)} \alpha^k_{ji} \mathbf{W}_u^k \mathbf{u}^{\prime}_i \right)  \,,  \\
& \mathbf{u}^{global}_i = \mathop{||}_{k=1}^{K} \sigma \left( \sum_{j \in \mathcal{N}(u_i)} \beta^k_{jt} \mathbf{W}_m^k \mathbf{m}^{\prime}_j  \right)    \,,
\end{split}
\label{final_global}
\end{equation}
where $||$ means concatenation operation and $\sigma(.)$ denotes the ELU~\cite{clevert2015fast} activation function.


\begin{algorithm}[!htb] 
    \caption{The global relation encoding algorithm.} 
    \small
    \label{alg_global} 
    \KwIn{
    	Global relation graph $G(V, E)$; user id $u_i$; microblog id $m_j$; weight vector $\mathbf{w}$; neighborhood function $\mathcal{N}(\cdot)$; $\textbf{Lookup}(\cdot)$ function transform user id to the user embedding. user behavior features $\mathbf{u}_f$, text representation $\mathbf{\widetilde{m}}$. 
    }
    
    \KwOut{
    	Microblog representations $\mathbf{m}^{global}_j$ and user representations $\mathbf{u}^{global}_i$, $\forall i, j \in V$. 
    }
    
    \For{$i, j \in V$}
    {
        \For{$m_j \in \mathcal{N}(u_i)$}
        {
    		$\mathbf{m}_j^{0} = \textbf{Lookup}(m_j)$ \;
    		
    		Calculate $\mathbf{u}^{\prime}_i$ and $\mathbf{m}^{\prime}_j$ by Equation~(\ref{global_representation}),~(\ref{transform_representation}) \;
    		
    		Calculate $\alpha_{ji}$ by Equation~(\ref{global_representation}),~(\ref{transform_representation}) \;
    		
        }
        Calculate $\mathbf{m}^{global}_j$ by Equation~(\ref{final_global}) \; 

        \For{$u_i \in \mathcal{N}(m_j)$}
        {
    		$\mathbf{u}_i^{0} = \textbf{Lookup}(u_i)$ \;
    		
    		Calculate $\mathbf{u}^{\prime}_i$ and $\mathbf{m}^{\prime}_j$ by Equation~(\ref{global_representation}),~(\ref{transform_representation}) \; 
    		
    		Calculate $\beta_{jt}$ by Equation~(\ref{global_representation}),~(\ref{transform_representation}) \;
    		
        }
        Calculate $\mathbf{u}^{global}_i$ by Equation~(\ref{final_global}) \;
    }
\end{algorithm}

More details are shown in Algorithm~\ref{alg_global}. The algorithm takes a graph as input and produces a latent representation of every node. At each iteration, every node collects node embedding from its local neighbors. Since these embeddings contribute differently to the representation of the current node, the attention mechanism is applied to obtain the relative weights of each node. Finally, the weighted average of neighbors' embeddings is treated as the representation of the current node. During the iteration process, the current node will gradually gain more and more information from multi-hops of the graph and finally obtain a global view of the graph.

\subsection{Rumor Classification}
After the above procedures, we get the local representation $\mathbf{\widetilde{m}}_j$ and the global representation $\mathbf{m}^{global}_j$. Both representations are important for rumor detection, thus they are concatenated as final features for classification. Then, fully-connected layers are applied to project the final representation into the target space of classes probability:
\begin{equation}
\begin{split}
& p_i(c| m_i, \mathcal{N}(m_i), \mathcal{U}; \theta) = \mathop{\textbf{softmax}}(\mathbf{W} [\mathbf{\widetilde{m}}_j; \mathbf{m}^{global}_j] + b)  \,,   \\
\end{split}
\end{equation}
where $\mathbf{W} \in \mathbb{R}^{2d \times |c|}$ is the weight parameter and $b \in \mathbb{R}$ is a bias term. 

Finally, the cross-entropy loss is used as the optimization objective function for rumor detection: 
\begin{equation}
	\begin{split}
	& J\left(c^{(i)}| D, u^{(i)}; \theta\right) = -\sum_{i}y_i\log p_i(c| m_i, \mathcal{N}(m_i), \mathcal{U}; \theta) \,,  \\
	\end{split}
\end{equation}
where $y_i$ is the gold probability of rumor class and $\theta$ represents all parameters of the model.



\section{Experiments} \label{experiments}

\subsection{Data sets}
We evaluate the proposed model on three real-world data collections: Weibo~\cite{ma2016detecting}, Twitter15~\cite{ma2017detect} and Twitter16~\cite{ma2017detect}, which were originally collected from the most popular social media website in China and the U.S. respectively. 

The Weibo data set only contains binary labels, i.e., ``false rumor'' and ``non-rumor''. The Twitter15 and Twitter16 data set each contains four different labels, i.e., ``false rumor'' (FR), ``non-rumor'' (NR), ``unverified'' (UR), and ``true rumor'' (TR). Note that the label ``true rumor'' denotes a microblog that tells people that a certain microblog is fake. For each data set, a heterogeneous graph is constructed from source tweets, responsive tweets, and related users. 
Table~\ref{tab1} shows the statistics of the three data sets. Since the original data sets do not contain user information, we crawled all the related user profiles via Twitter API\footnote{https://dev.twitter.com/rest/public}.

\begin{table}[htbp]
\centering
\caption{data set statistics.}
	\setlength{\tabcolsep}{2.5mm}{
        \begin{tabular}{c|c c c}
            \hline
            \textbf{Statistic}&\textbf{Weibo} &\textbf{Twitter15} &\textbf{Twitter16} \\
            \hline
            \hline
            \# source tweets& 4664 & 1490 &818 \\
            \hline
            \# non-rumors& 2351 & 374 &205 \\
            \hline
            \# false rumors& 2313 & 370 &205 \\
            \hline
            \# unverified rumors& 0 & 374 &203 \\
            \hline
            \# true rumors& 0 & 372 &205 \\
            \hline
            \# users& 2,746,818 & 276,663 &173,487 \\
            \hline
            \# posts& 3,805,656 & 331,612 &204,820 \\
            \hline
        \end{tabular}
    }
\label{tab1}
\end{table}

\begin{table*}[!htbp]
	\begin{minipage}[t]{0.5\linewidth}
		\centering
		\caption{Experimental results on Twitter15 data set. }
		\setlength{\tabcolsep}{3mm}{
			\begin{tabular}{|c|c|cccc|}
			    \hline
                \multicolumn{6}{|c|}{\textit{Twitter15}} \\ \hline
                \multirow{2}{*}{Method} & \multirow{2}{*}{Acc} & NR & FR & TR & UR \\ \cline{3-6}
                 &  & $F_1$ & $F_1$ & $F_1$ & $F_1$ \\
                \hline
                DTR    & 0.409     & 0.501     & 0.311 & 0.364 & 0.473  \\
                \hline
                DTC    & 0.454     & 0.733     & 0.355 & 0.317 & 0.415  \\
                \hline
                RFC    & 0.565     & 0.810     & 0.422 & 0.401 & 0.543  \\
                \hline
                SVM-RBF    & 0.318     & 0.455     & 0.037 & 0.218 & 0.225  \\
                \hline
                SVM-TS    & 0.544     & 0.796     & 0.472 & 0.404 & 0.483  \\
                \hline
                PTK    & 0.750     & 0.804     & 0.698 & 0.765 & 0.733  \\
                \hline
                GRU    & 0.646     & 0.792     & 0.574 & 0.608 & 0.592  \\
                \hline
                RvNN    & 0.723     & 0.682     & 0.758 & 0.821 & 0.654  \\
                \hline
                PPC    & 0.842     & 0.811     & 0.875 & 0.818 & 0.790  \\
                \hline
                GLAN    & \textbf{0.905}     & \textbf{0.924}     & \textbf{0.917} & \textbf{0.852} & \textbf{0.927}  \\
                \hline
			\end{tabular}
			\label{exp_results_on_twitter15}
		}
	\end{minipage} 
	\begin{minipage}[t]{0.5\linewidth}  
	    \centering
		\caption{
			Experimental results  on Twitter16 data set. 
		}
	    \setlength{\tabcolsep}{3mm}{
	    	\begin{tabular}{|c|c|cccc|}
	    	        \hline
	    	        \multicolumn{6}{|c|}{\textit{Twitter16}} \\ \hline
                    \multirow{2}{*}{Method} & \multirow{2}{*}{Acc} & NR & FR & TR & UR \\ \cline{3-6}
                     &  & $F_1$ & $F_1$ & $F_1$ & $F_1$ \\
                    
                    \hline
                    DTR    & 0.414     & 0.394     & 0.273 & 0.630 & 0.344  \\
                    \hline
                    DTC    & 0.465     & 0.643 & 0.393…& 0.419 & 0.403  \\
                    \hline
                    RFC    & 0.585     & 0.752 & 0.415 & 0.547 & 0.563  \\
                    \hline
                    SVM-RBF    & 0.321     & 0.423 & 0.085 & 0.419 & 0.037 \\
                    \hline
                    SVM-TS    & 0.574     & 0.755 & 0.420 & 0.571 & 0.526  \\
                    \hline
                    PTK    & 0.732     & 0.740 & 0.709 & 0.836 & 0.686  \\
                    \hline
                    GRU    & 0.633     & 0.772 & 0.489 & 0.686 & 0.593  \\
                    \hline
                    RvNN    & 0.737     & 0.662     & 0.743 & 0.835 & 0.708  \\
                    \hline
                    PPC    & 0.863     & 0.820 & 0.898 & 0.843 & 0.837  \\
                    \hline
                    GLAN    & \textbf{0.902}     & \textbf{0.921}     & 0.869 & \textbf{0.847} & \textbf{0.968}  \\
                    \hline
	    	\end{tabular}
	    	\label{exp_results_on_twitter16}
	    }
	\end{minipage}
\end{table*}

\begin{table}[h]
    \centering
    \caption{Rumor detection results on the weibo data set.}
    \setlength{\tabcolsep}{3mm}{
        \begin{tabular}{|c|c|cccc|}
            \hline
            Method                  & Class & Acc. & Prec. & Recall & $F_1$ \\
            
            \hline
            \multirow{2}{*}{DTR}    & FR     & \multirow{2}{*}{0.732}      & 0.738 & 0.715 & 0.726  \\
                                    & NR     &                         & 0.726 & 0.749 & 0.737  \\
            \hline
            \multirow{2}{*}{DTC}    & FR     & \multirow{2}{*}{0.831}      & 0.847 & 0.815 & 0.831  \\
                                    & NR     &                         & 0.815 & 0.847 & 0.830  \\
            \hline
            \multirow{2}{*}{RFC}    & FR     & \multirow{2}{*}{0.849}      & 0.786 & 0.959 & 0.864  \\
                                    & NR     &                         & 0.947 & 0.739 & 0.830  \\
            \hline
            \multirow{2}{*}{SVM-RBF} & FR     & \multirow{2}{*}{0.818}      & 0.822 & 0.812 & 0.817  \\
                                    & NR     &                         & 0.815 & 0.824 & 0.819  \\
            \hline
            \multirow{2}{*}{SVM-TS} & FR     & \multirow{2}{*}{0.857}      & 0.839 & 0.885 & 0.861  \\
                                    & NR     &                         & 0.878 & 0.830 & 0.857  \\
            \hline
            \multirow{2}{*}{GRU}    & FR     & \multirow{2}{*}{0.910}      & 0.876 & 0.956 & 0.914  \\
                                    & NR     &                         & \textbf{0.952} & 0.864 & 0.906  \\
            \hline
            \multirow{2}{*}{PPC}    & FR     & \multirow{2}{*}{0.921}      & 0.896 & \textbf{0.962} & 0.923  \\
                                    & NR     &                         & 0.949 & 0.889 & 0.918  \\
            \hline
            \multirow{2}{*}{GLAN}    & FR     & \multirow{2}{*}{\textbf{0.946}}      & \textbf{0.943} & 0.948 & \textbf{0.945}  \\
                                    & NR     &                         & 0.949 & \textbf{0.943} & \textbf{0.946}  \\
            \hline
        \end{tabular}
    }
    \label{exp_results_on_weibo}
\end{table}

\subsection{Baseline Models}
We compare our model with a series of baseline rumor detection models as follows:
\begin{itemize}
	\item \textbf{DTC}~\cite{Castillo_2011}: A decision tree-based model that utilizes a combination of news characteristics.
	\item \textbf{SVM-RBF}~\cite{Yang_2012}: An SVM model with RBF kernel that utilize a combination of news characteristics.
	\item \textbf{SVM-TS}~\cite{Ma_2015}: An SVM model that utilizes time-series to model the variation of news characteristics.
	\item \textbf{DTR}~\cite{Zhao_2015}: A decision-tree-based ranking method for detecting fake news through enquiry phrases.
	\item \textbf{GRU}~\cite{ma2016detecting}: A RNN-based model that learns temporal-linguistic patterns from user comments.
	\item \textbf{RFC}~\cite{Kwon_2017}: A random forest classifier that utilizes user, linguistic and structure characteristics.
	\item \textbf{PTK}~\cite{ma2017detect}: An SVM classifier with a propagation tree kernel that detects fake news by learning temporal-structure patterns from propagation trees.
	\item \textbf{RvNN}~\cite{ma2018rumor}: A bottom-up and a top-down tree-structured model based on recursive neural networks for rumor detection on Twitter.
	\item \textbf{PPC}~\cite{liu2018early}: A novel model that detects fake news through propagation path classification with a combination of recurrent and convolutional networks.
\end{itemize}
The \textbf{PPC} model is the state-of-the-art method for rumor detection when submitting this paper. 

\subsection{Data Preprocessing}
Same as the original papers~\cite{ma2016detecting,ma2018rumor}, we randomly select 10\% instances as the development data set, and split the rest for training and testing set with a ratio of 3:1 in all three data sets. 

All word embeddings in the model are initialized with the 300-dimensional word vectors, which are trained on domain-specific review corpora by Skip-gram~\cite{mikolov2013distributed} algorithm. Words that are not present at the set of pre-trained word vectors are initialized from a uniform distribution. We keep the word vectors trainable in the training process, and they can be fine-tuned for each task.

For Twitter15 and Twitter16 data set, the words are segmented by white space. And for the Weibo data set, the words are segmented by \textit{Jieba}\footnote{https://github.com/fxsjy/jieba} library. We remove words with less than 2 occurrences because they may be stop words.

\subsection{Evaluation Metrics and Parameter Settings} 
For fair comparison, we adopt the same evaluation metrics used in the prior work~\cite{liu2018early,ma2018rumor}. Thus, the accuracy, precision, recall and F1 score are adopted for evaluation. 

Our model is implemented by PyTorch\footnote{Our code and data will be
available at https://github.com/chunyuanY/RumorDetection}~\cite{paszke2017automatic}. The parameters are updated by Adam algorithm~\cite{kingma2014adam} and the parameters of Adam, $\beta_1$ and $\beta_2$ are 0.9 and 0.999 respectively. The learning rate is initialized as 1e-3 and gradually decreased during the process of training. We select the best parameter configuration based on performance on the development set and evaluate the configuration on the test set. The convolutional kernel size is set to (3, 4, 5) with 100 kernels for each kind of size. The number of heads K is set to 8. The batch size of the training set is set to 64.

\subsection{Results and Analysis}
Table~\ref{exp_results_on_weibo}, \ref{exp_results_on_twitter15} and \ref{exp_results_on_twitter16} show the performance of all compared methods. For fair comparison, the experimental results of baseline models are directly cited from previous studies~\cite{ma2018rumor,liu2018early}. We also bold the best result of each column in all tables. 

From the Table~\ref{exp_results_on_weibo}, \ref{exp_results_on_twitter15} and \ref{exp_results_on_twitter16}, we can observe that the GLAN outperforms all other baselines on three data sets. Specifically, our model achieves an accuracy of 94.6\% on Weibo data set and 90.5\%, 90.2\% on two Twitter data sets, which indicates the flexibility of our model on different types of data sets. Moreover, the outstanding results indicate that the heterogeneous graph models with local and global attention can effectively learn the representation of nodes using semantic and structural information.

It is observed that the performance of the methods based on hand-crafted features (DTR, DTC, RFC, SVM-RBF, and SVM-TS) is obviously poor, indicating that they fail to generalize due to the lack of capacity capturing helpful features. Among these baselines, SVM-TS and RFC perform relatively better because they use additional temporal or structural features, but they are still clearly worse than the models not relying on feature engineering. 

Among the two propagation tree-based methods, PTK relies on both linguistic and structural features extracted from propagation trees. RvNN model is inherently tree-structured and takes advantage of representation learning following the propagation structure, thus beats PTK. However, these tree-based methods will lose too much information during modeling the propagation process, because the messages are spread by a graph structure rather than a tree structure. 

For deep learning methods, GRU and PPC outperform traditional classifiers that using manually crafted features. This observation indicates that the neural network model can learn deep latent features automatically. We also observe that PPC is much more effective than GRU. There are two reasons: 1) GRU relies on temporal-linguistic patterns, while PPC relies on the fixed user characteristics of repost sequences. 2) PPC combines both CNN and RNN to capture the variations of user characteristics.

\begin{figure*}[!htbp]
	\centering 
	\subfigure[Weibo]{
		\label{fig:subfig:c1} 
		\includegraphics[scale=0.63]{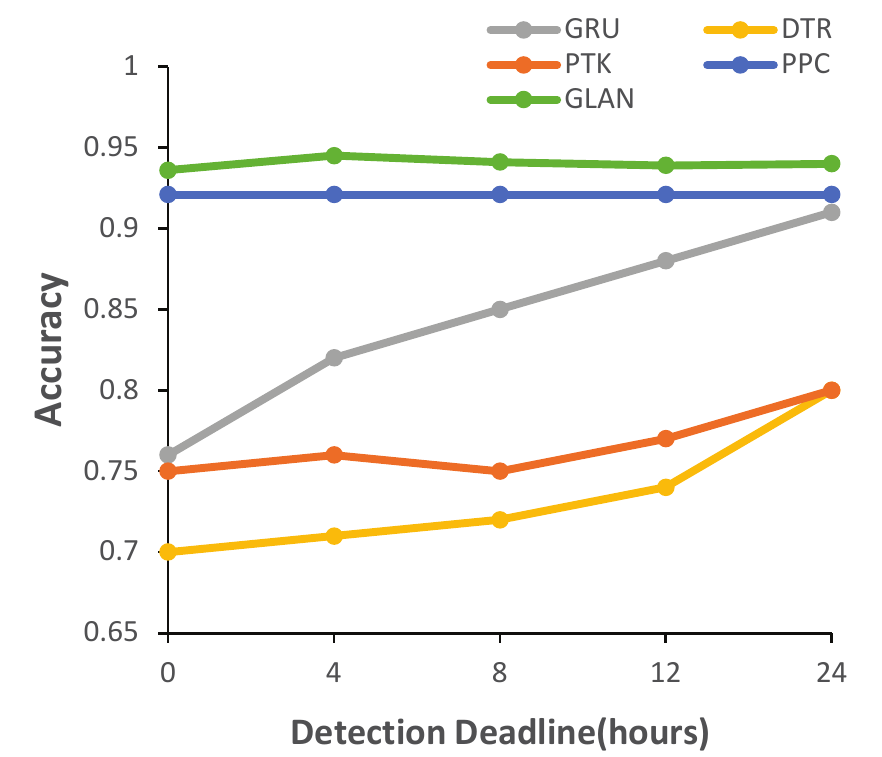} 
	}
	\subfigure[Twitter16]{
		\label{fig:subfig:a1} 
		\includegraphics[scale=0.63]{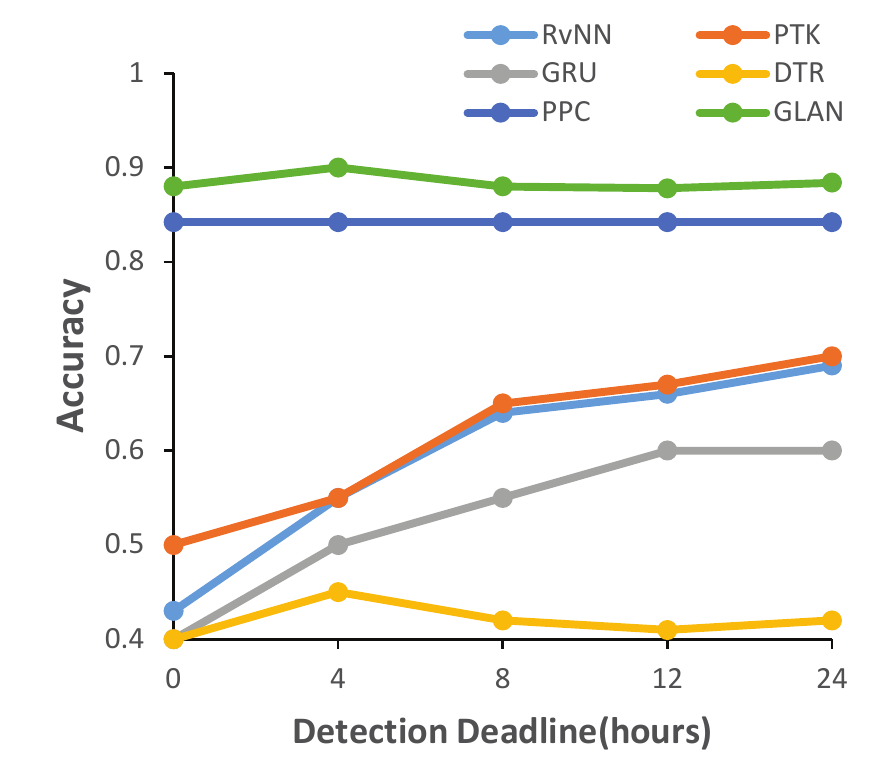} 
	}
	\subfigure[Twitter15]{
		\label{fig:subfig:b1} 
		\includegraphics[scale=0.63]{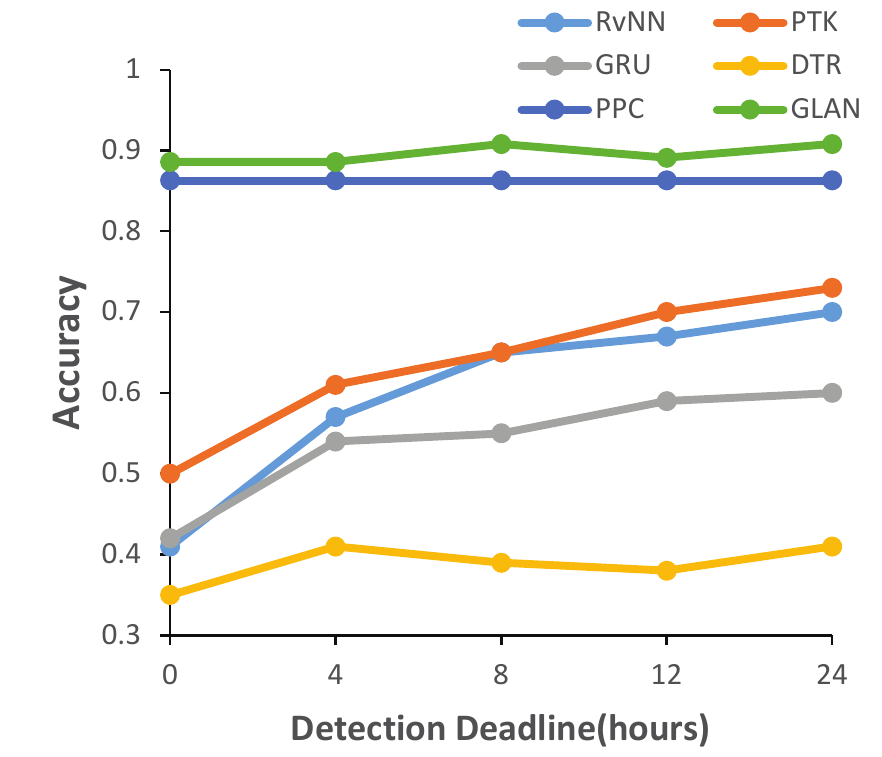} 
	}
    \caption{Results of early rumor detection.}
    \label{early_detection}
\end{figure*}

In conclusion, the GLAN outperforms neural network-based models and features based methods, and the precision and accuracy of rumor detection obtain significant improvement. Specifically, on the Weibo data set, our model increases the accuracy of the propagation path classification method (the best baseline) from 92.1\% to 94.6\%. On the Twitter data sets, the accuracy is boosted from 84.2\% to 90.5\% and 86.3\% to 90.2\%, respectively. This results demonstrate that the local semantic information and global structural information are critical for learning the differences between rumors and non-rumors.

\subsection{Ablation Study}
In order to determine the relative importance of every module of the GLAN, we perform a series of ablation studies over the different parts of the model. The experimental results are presented in Table~\ref{ablation_results}. The ablation studies are conducted as following orders: 
\begin{itemize}
	\item \textbf{w/o LRE}: Removing local relation encoding module and only using global structure information for rumor classification.  
	\item \textbf{w/o GRE}: Removing global relation encoding module and only remaining local structure encoding module for rumor classification. 
	\item \textbf{Only Text}: Only using CNN to extract text features from source tweet for rumor classification.
\end{itemize}

\begin{table}[!htb]
	\centering
	\caption{
		The ablation study results on the Weibo, Twitter15, and Twitter16 data set. 
	}
	\setlength{\tabcolsep}{3mm}{
		\begin{tabular}{p{2cm}|c|c|c}
			\toprule
			\multirow{2}[3]{*}{Models} & Weibo & Twitter15 & Twitter16 \\
			& Accuracy & Accuracy & Accuracy \\
			\midrule
			GLAN            &94.6  &90.5  &90.2    \\
			w/o LRE         &86.8  &82.7  &86.4    \\
			w/o GRE         &88.7  &83.8  &87.5    \\
			Only Text       &81.4  &75.4  &71.3    \\
			\bottomrule
		\end{tabular}
	}
	\label{ablation_results}
\end{table}

\noindent From experimental results in Table~\ref{ablation_results}, we can observe that:

We first examine the impact brought by the local relation encoding module. We can see that removing the LRE significantly affects performance on all data sets, and the accuracy drops 7.8\%, 7.8\%, and 3.8\% on the Weibo, Twitter15, and Twitter16 data sets. LRE captures the semantic relation between source tweet and corresponding retweet, and thus the results show it is very important to explicitly encode the local relation. 

Next, we evaluate the influence of the global relation encoding module. Referring to Table~\ref{ablation_results}, the lack of GRE still brings significant performances decline on all data sets. Intuitively speaking, the GRE module will make the distance between two users with identical class labels close to each other, thus leading to the rumor and non-rumor group high cohesion and low coupling, and thereby it can improve the performance. 

The performance obtains significant improvement after combining local and global relation information, which shows the combination of two aspect information provide complementary effect from the local and global aspect.

\begin{figure*}[!htbp]
	\centering 
	\subfigure[Tweet length]{
		\label{tweet_length} 
		\includegraphics[scale=0.95]{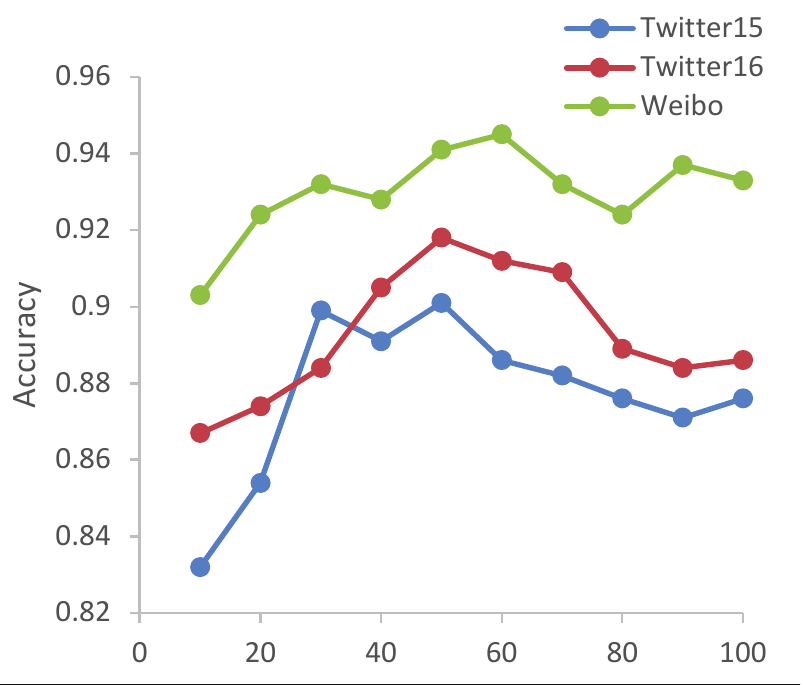} 
	}
	\subfigure[Convolutional kernel sizes]{
		\label{kernel_sizes} 
		\includegraphics[scale=0.95]{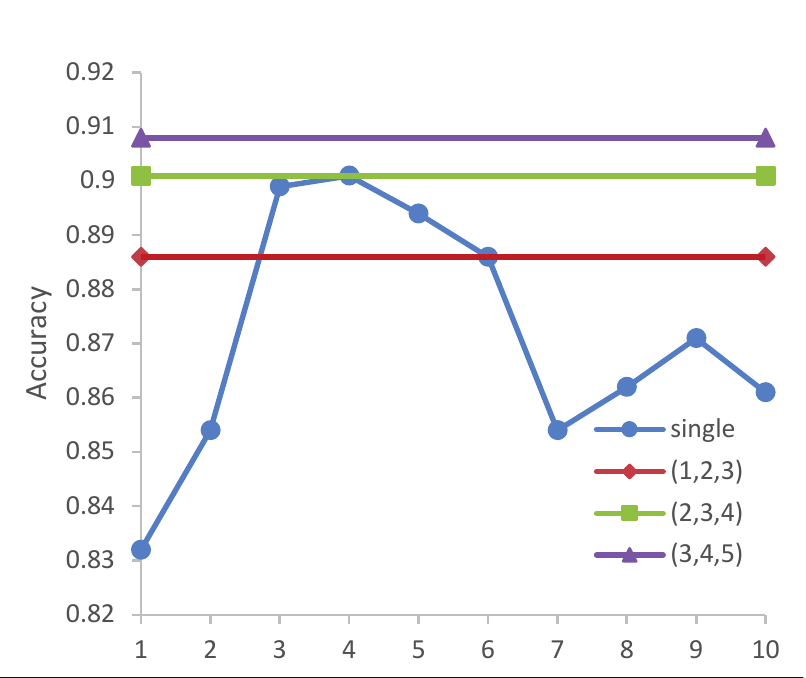} 
	}
	\caption{Effect of different (a) tweet length and (b) convolutional kernel sizes on the data sets. ``Single'': means single fixed kernel size, and some multi-sizes combinations are explored in experiments.}
\end{figure*}

\subsection{Early Detection}
In rumor detection, one of the most crucial goals is to detect rumors as early as possible so that interventions can be made in time~\cite{Zhao_2015}. By setting a detection delay time, only tweets posted before than the delay can be used for evaluating early detection performance. We compared different methods in terms of different time delays, and the performance is evaluated by the accuracy obtained when we incrementally add test data up to the checkpoint given the targeted time delay. 

By varying the time delays of retweets, the accuracy of several competitive models is shown in Fig.~\ref{early_detection}. In the first few hours, our model GLAN using less than 4-hour data has already outperformed the tree-based classification methods (the best baseline) using all-time data, indicating the superior early detection performance of our model. Particularly, GLAN achieves 94\% accuracy on Weibo, 88\% accuracy on Twitter15 and 90\% on Twitter16 within 4 hours, which is much faster than other models. 

When the time delays vary from 4 to 12 hours, our model has a slight drop, but it still works better than the state-of-the-art models. This is because With the propagation of microblog there is more structural and semantic information, while the noise is increased at the same time. Therefore, the results show that our model is insensitive to data and has better stability and robustness. 

Experimental results on three real-world data sets demonstrate that the proposed model can significantly improve detection performance and enhance the effectiveness of early detection of rumors at the same time.

\section{Parameter Analysis} \label{parameter_analysis}
The GLAN utilizes the fixed size of convolutional kernels and microblog text length. So the selection of these hyper-parameters may influence the performance of the model. In this section, we will conduct a series of parameter sensitivity experiments to study the effect of these hyper-parameters. In the experiments, we apply the control variates method to reduce the influence of irrelevant parameters.

Figure~\ref{tweet_length} demonstrates the performance with different tweet length. It is obvious that the tweet length has a significant influence on classification performance. When the tweet is too short, it cannot provide enough information for classification. So the performance gets improvement as the growth of tweet length. 

Figure~\ref{kernel_sizes} indicates the performance with different kernel sizes. When the kernel size is set to 1, it captures uni-gram features that are similar to a uni-gram bag of words. Intuitively, uni-gram features will totally lose the word location information thus influencing classification performance. The experiments on three data sets show that the performances will increase with the growth of kernel size, and the peak is 3 or 4. Moreover, we conduct another experiment with the combination of multiple different convolutional kernel sizes. The experiment results demonstrate that the combination of multiple kernel sizes (3, 4, 5) acquires the best performance. The effectiveness of multiple kernel combinations can be explained that different kernel sizes have different respective fields to capture more distinctive phrases.

\section{Conclusion and Future Work} \label{conclusion}
In this paper, we propose a heterogeneous graph with local and global attention, which combines the local semantic and global structural information for rumor detection. Different from most existing studies extracting hand-crafted features or feeding repost sequences to network directly, we fuse the local contextual information from source tweet and corresponding retweets via multi-head attention and generate a better integrated representation for each source tweet. To capture complex global information from different source tweets, we build a heterogeneous graph using global structural information with global attention to detect rumors. Extensive experiments conducted on Weibo and Twitter data sets show that the proposed model can significantly outperform other state-of-the-art models on both rumor classification and  early detection task.

In the future work, we plan to integrate other types of information such as user profiles and geographic location into the local heterogeneous graphs. Moreover, we will explore more efficient methods to build a global graph to utilize new structural information for further enhancing the text representation learning and detect rumor spreaders at an early time.

\section{Acknowledge}
We thank the anonymous reviewers for their feedback. This research is supported in part by the Beijing Municipal Science and Technology Project under Grant Z191100007119008 and Z181100002718004, the National Key Research and Development Program of China under Grant 2018YFC0806900 and 2017YFB1010000. 

%
\bibliographystyle{./IEEEtran}
\bibliography{./icdm.bib}

\begin{thebibliography}{10}
\providecommand{\url}[1]{#1}
\csname url@samestyle\endcsname
\providecommand{\newblock}{\relax}
\providecommand{\bibinfo}[2]{#2}
\providecommand{\BIBentrySTDinterwordspacing}{\spaceskip=0pt\relax}
\providecommand{\BIBentryALTinterwordstretchfactor}{4}
\providecommand{\BIBentryALTinterwordspacing}{\spaceskip=\fontdimen2\font plus
\BIBentryALTinterwordstretchfactor\fontdimen3\font minus
  \fontdimen4\font\relax}
\providecommand{\BIBforeignlanguage}[2]{{%
\expandafter\ifx\csname l@#1\endcsname\relax
\typeout{** WARNING: IEEEtran.bst: No hyphenation pattern has been}%
\typeout{** loaded for the language `#1'. Using the pattern for}%
\typeout{** the default language instead.}%
\else
\language=\csname l@#1\endcsname
\fi
#2}}
\providecommand{\BIBdecl}{\relax}
\BIBdecl

\bibitem{Castillo_2011}
C.~Castillo, M.~Mendoza, and B.~Poblete, ``Information credibility on
  twitter,'' in \emph{Proceedings of the 20th international conference on World
  wide web}.\hskip 1em plus 0.5em minus 0.4em\relax ACM, 2011, pp. 675--684.

\bibitem{Qazvinian_2011}
\BIBentryALTinterwordspacing
V.~Qazvinian, E.~Rosengren, D.~R. Radev, and Q.~Mei, ``Rumor has it:
  Identifying misinformation in microblogs,'' in \emph{Proceedings of the
  Conference on Empirical Methods in Natural Language Processing}, ser. EMNLP
  '11.\hskip 1em plus 0.5em minus 0.4em\relax Stroudsburg, PA, USA: Association
  for Computational Linguistics, 2011, pp. 1589--1599. [Online]. Available:
  \url{http://dl.acm.org/citation.cfm?id=2145432.2145602}
\BIBentrySTDinterwordspacing

\bibitem{Popat_2017}
\BIBentryALTinterwordspacing
K.~Popat, ``Assessing the credibility of claims on the web,'' in
  \emph{Proceedings of the 26th International Conference on World Wide Web
  Companion}, ser. WWW '17 Companion.\hskip 1em plus 0.5em minus 0.4em\relax
  Republic and Canton of Geneva, Switzerland: International World Wide Web
  Conferences Steering Committee, 2017, pp. 735--739. [Online]. Available:
  \url{https://doi.org/10.1145/3041021.3053379}
\BIBentrySTDinterwordspacing

\bibitem{Yang_2012}
\BIBentryALTinterwordspacing
F.~Yang, Y.~Liu, X.~Yu, and M.~Yang, ``Automatic detection of rumor on sina
  weibo,'' \emph{Proceedings of the ACM SIGKDD Workshop on Mining Data
  Semantics - MDS ’12}, 2012. [Online]. Available:
  \url{http://dx.doi.org/10.1145/2350190.2350203}
\BIBentrySTDinterwordspacing

\bibitem{jin2013epidemiological}
F.~Jin, E.~Dougherty, P.~Saraf, Y.~Cao, and N.~Ramakrishnan, ``Epidemiological
  modeling of news and rumors on twitter,'' in \emph{Proceedings of the 7th
  Workshop on Social Network Mining and Analysis}.\hskip 1em plus 0.5em minus
  0.4em\relax ACM, 2013, p.~8.

\bibitem{sampson2016leveraging}
J.~Sampson, F.~Morstatter, L.~Wu, and H.~Liu, ``Leveraging the implicit
  structure within social media for emergent rumor detection,'' in
  \emph{Proceedings of the 25th ACM International on Conference on Information
  and Knowledge Management}.\hskip 1em plus 0.5em minus 0.4em\relax ACM, 2016,
  pp. 2377--2382.

\bibitem{ma2017detect}
J.~Ma, W.~Gao, and K.-F. Wong, ``Detect rumors in microblog posts using
  propagation structure via kernel learning,'' in \emph{Proceedings of the 55th
  Annual Meeting of the Association for Computational Linguistics (Volume 1:
  Long Papers)}, 2017, pp. 708--717.

\bibitem{liu2018early}
Y.~Liu and Y.-F.~B. Wu, ``Early detection of fake news on social media through
  propagation path classification with recurrent and convolutional networks,''
  in \emph{Thirty-Second AAAI Conference on Artificial Intelligence}, 2018.

\bibitem{song2018ced}
C.~Song, C.~Tu, C.~Yang, Z.~Liu, and M.~Sun, ``Ced: Credible early detection of
  social media rumors,'' \emph{arXiv preprint arXiv:1811.04175}, 2018.

\bibitem{ma2016detecting}
J.~Ma, W.~Gao, P.~Mitra, S.~Kwon, B.~J. Jansen, K.-F. Wong, and M.~Cha,
  ``Detecting rumors from microblogs with recurrent neural networks.'' in
  \emph{Ijcai}, 2016, pp. 3818--3824.

\bibitem{tang2015line}
J.~Tang, M.~Qu, M.~Wang, M.~Zhang, J.~Yan, and Q.~Mei, ``Line: Large-scale
  information network embedding.'' in \emph{WWW}.\hskip 1em plus 0.5em minus
  0.4em\relax ACM, 2015.

\bibitem{Ma_2015}
\BIBentryALTinterwordspacing
J.~Ma, W.~Gao, Z.~Wei, Y.~Lu, and K.-F. Wong, ``Detect rumors using time series
  of social context information on microblogging websites,'' \emph{Proceedings
  of the 24th ACM International on Conference on Information and Knowledge
  Management - CIKM ’15}, 2015. [Online]. Available:
  \url{http://dx.doi.org/10.1145/2806416.2806607}
\BIBentrySTDinterwordspacing

\bibitem{Kwon_2013}
\BIBentryALTinterwordspacing
S.~Kwon, M.~Cha, K.~Jung, W.~Chen, and Y.~Wang, ``Prominent features of rumor
  propagation in online social media,'' \emph{2013 IEEE 13th International
  Conference on Data Mining}, Dec 2013. [Online]. Available:
  \url{http://dx.doi.org/10.1109/ICDM.2013.61}
\BIBentrySTDinterwordspacing

\bibitem{Yu_2017}
\BIBentryALTinterwordspacing
F.~Yu, Q.~Liu, S.~Wu, L.~Wang, and T.~Tan, ``A convolutional approach for
  misinformation identification,'' \emph{Proceedings of the Twenty-Sixth
  International Joint Conference on Artificial Intelligence}, Aug 2017.
  [Online]. Available: \url{http://dx.doi.org/10.24963/ijcai.2017/545}
\BIBentrySTDinterwordspacing

\bibitem{Ruchansky_2017}
\BIBentryALTinterwordspacing
N.~Ruchansky, S.~Seo, and Y.~Liu, ``Csi: A hybrid deep model for fake news
  detection,'' \emph{Proceedings of the 2017 ACM on Conference on Information
  and Knowledge Management - CIKM ’17}, 2017. [Online]. Available:
  \url{http://dx.doi.org/10.1145/3132847.3132877}
\BIBentrySTDinterwordspacing

\bibitem{Bhatt_2018}
\BIBentryALTinterwordspacing
G.~Bhatt, A.~Sharma, S.~Sharma, A.~Nagpal, B.~Raman, and A.~Mittal, ``Combining
  neural, statistical and external features for fake news stance
  identification,'' \emph{Companion of the The Web Conference 2018 on The Web
  Conference 2018 - WWW ’18}, 2018. [Online]. Available:
  \url{http://dx.doi.org/10.1145/3184558.3191577}
\BIBentrySTDinterwordspacing

\bibitem{Wu_2015}
\BIBentryALTinterwordspacing
K.~Wu, S.~Yang, and K.~Q. Zhu, ``False rumors detection on sina weibo by
  propagation structures,'' \emph{2015 IEEE 31st International Conference on
  Data Engineering}, Apr 2015. [Online]. Available:
  \url{http://dx.doi.org/10.1109/ICDE.2015.7113322}
\BIBentrySTDinterwordspacing

\bibitem{kim2014convolutional}
Y.~Kim, ``Convolutional neural networks for sentence classification,'' in
  \emph{Proceedings of the 2014 Conference on Empirical Methods in Natural
  Language Processing (EMNLP)}, 2014, pp. 1746--1751.

\bibitem{kalchbrenner2014convolutional}
N.~Kalchbrenner, E.~Grefenstette, and P.~Blunsom, ``A convolutional neural
  network for modelling sentences,'' in \emph{Proceedings of the 52nd Annual
  Meeting of the Association for Computational Linguistics (Volume 1: Long
  Papers)}, vol.~1, 2014, pp. 655--665.

\bibitem{tai2015improved}
K.~S. Tai, R.~Socher, and C.~D. Manning, ``Improved semantic representations
  from tree-structured long short-term memory networks,'' in \emph{Proceedings
  of the 53rd Annual Meeting of the Association for Computational Linguistics},
  2015.

\bibitem{yang2016hierarchical}
Z.~Yang, D.~Yang, C.~Dyer, X.~He, A.~Smola, and E.~Hovy, ``Hierarchical
  attention networks for document classification,'' in \emph{Proceedings of the
  2016 Conference of the North American Chapter of the Association for
  Computational Linguistics: Human Language Technologies}, 2016, pp.
  1480--1489.

\bibitem{ma2018rumor}
J.~Ma, W.~Gao, and K.-F. Wong, ``Rumor detection on twitter with
  tree-structured recursive neural networks,'' in \emph{Proceedings of the 56th
  Annual Meeting of the Association for Computational Linguistics (Volume 1:
  Long Papers)}, 2018, pp. 1980--1989.

\bibitem{vaswani2017attention}
A.~Vaswani, N.~Shazeer, N.~Parmar, J.~Uszkoreit, L.~Jones, A.~N. Gomez,
  {\L}.~Kaiser, and I.~Polosukhin, ``Attention is all you need,'' in
  \emph{Advances in neural information processing systems}, 2017, pp.
  5998--6008.

\bibitem{kipf2017semi}
\BIBentryALTinterwordspacing
T.~N. Kipf and M.~Welling, ``Semi-supervised classification with graph
  convolutional networks,'' in \emph{International Conference on Learning
  Representations}, 2017. [Online]. Available:
  \url{https://openreview.net/forum?id=SJU4ayYgl}
\BIBentrySTDinterwordspacing

\bibitem{hamilton2017inductive}
W.~Hamilton, Z.~Ying, and J.~Leskovec, ``Inductive representation learning on
  large graphs,'' in \emph{Advances in Neural Information Processing Systems},
  2017, pp. 1024--1034.

\bibitem{veli2018graph}
\BIBentryALTinterwordspacing
P.~Veličković, G.~Cucurull, A.~Casanova, A.~Romero, P.~Liò, and Y.~Bengio,
  ``Graph attention networks,'' in \emph{International Conference on Learning
  Representations}, 2018. [Online]. Available:
  \url{https://openreview.net/forum?id=rJXMpikCZ}
\BIBentrySTDinterwordspacing

\bibitem{xu2015empirical}
B.~Xu, N.~Wang, T.~Chen, and M.~Li, ``Empirical evaluation of rectified
  activations in convolutional network,'' \emph{arXiv preprint
  arXiv:1505.00853}, 2015.

\bibitem{clevert2015fast}
D.-A. Clevert, T.~Unterthiner, and S.~Hochreiter, ``Fast and accurate deep
  network learning by exponential linear units (elus),'' \emph{arXiv preprint
  arXiv:1511.07289}, 2015.

\bibitem{Zhao_2015}
Z.~Zhao, P.~Resnick, and Q.~Mei, ``Enquiring minds: Early detection of rumors
  in social media from enquiry posts,'' in \emph{Proceedings of the 24th
  International Conference on World Wide Web}, ser. WWW'15.\hskip 1em plus
  0.5em minus 0.4em\relax Republic and Canton of Geneva, Switzerland:
  International World Wide Web Conferences Steering Committee, 2015, pp.
  1395--1405.

\bibitem{Kwon_2017}
\BIBentryALTinterwordspacing
S.~Kwon, M.~Cha, and K.~Jung, ``Rumor detection over varying time windows,''
  \emph{PLOS ONE}, vol.~12, no.~1, pp. 1--19, 01 2017. [Online]. Available:
  \url{https://doi.org/10.1371/journal.pone.0168344}
\BIBentrySTDinterwordspacing

\bibitem{mikolov2013distributed}
T.~Mikolov, I.~Sutskever, K.~Chen, G.~S. Corrado, and J.~Dean, ``Distributed
  representations of words and phrases and their compositionality,'' in
  \emph{Advances in neural information processing systems}, 2013, pp.
  3111--3119.

\bibitem{paszke2017automatic}
A.~Paszke, S.~Gross, S.~Chintala, G.~Chanan, E.~Yang, Z.~DeVito, Z.~Lin,
  A.~Desmaison, L.~Antiga, and A.~Lerer, ``Automatic differentiation in
  pytorch,'' in \emph{NIPS-W}, 2017.

\bibitem{kingma2014adam}
D.~P. Kingma and J.~Ba, ``Adam: A method for stochastic optimization,''
  \emph{arXiv preprint arXiv:1412.6980}, 2014.

\end{thebibliography}

\end{document}